\pdfoutput=1
\PassOptionsToPackage{table}{xcolor}
\documentclass[11pt, a4paper, nonumbering]{deepseek}

\usepackage[T1]{fontenc}
\usepackage[utf8]{inputenc}
\usepackage{microtype}
\usepackage{graphicx}
\usepackage{booktabs}
\usepackage{array}
\usepackage{multirow}
\usepackage{amsfonts}
\usepackage{amsmath}
\usepackage{amssymb}
\usepackage{mathtools}
\usepackage{bm}
\usepackage{xspace}
\usepackage{pifont}
\usepackage{enumitem}
\usepackage{caption}
\usepackage[table]{xcolor}
\usepackage{colortbl}
\usepackage{titlesec}
\usepackage{xltabular}
\usepackage{longtable}
\usepackage{algorithm}
\usepackage{algorithmic}
\microtypesetup{patch=none}
\usepackage[bottom,flushmargin]{footmisc}
\usepackage{setspace}
\usepackage[authoryear, sort&compress, round]{natbib}
\definecolor{sbEvA}   {HTML}{2A78D6}   % event 1, Wuhan Library
\definecolor{sbEvB}   {HTML}{EB6834}   % event 2, US-Iran
\definecolor{sbEvC}   {HTML}{1BAF7A}   % event 3, TikTok
\definecolor{sbEvD}   {HTML}{EDA100}   % event 4, SMCI
\definecolor{sbEvE}   {HTML}{E87BA4}   % event 5, Trump Tariff
\definecolor{sbCal}   {HTML}{2A78D6}   % probability-calibration axis
\definecolor{sbTime}  {HTML}{EB6834}   % temporal-accuracy axis
\definecolor{sbBoth}  {HTML}{1BAF7A}   % the two axes combined
\definecolor{sbAnchor}{HTML}{4A3AA7}   % trivial anchor of 50 (reference, not a series)
\definecolor{sbTruth} {HTML}{D03B3B}   % largest gap / status-critical
\definecolor{sbHead}  {HTML}{184F95}   % headings + links (blue ramp, step 600)
\definecolor{sbTint}  {HTML}{EDF3FD}   % table header wash
\definecolor{sbTintB} {HTML}{F5F4F1}   % neutral wash for baseline blocks
\definecolor{sbRule}  {HTML}{9FB4CE}
\definecolor{sbInk2}  {HTML}{52514E}

\hypersetup{colorlinks=true,linkcolor=sbHead,citecolor=sbHead,urlcolor=sbHead}
\usepackage[capitalize,noabbrev]{cleveref}

\titleformat{\section}{\large\bfseries\headingfont\color{sbHead}}{\color{sbHead}\thesection.}{0.5em}{#1}[]
\titleformat{name=\section,numberless}{\large\bfseries\headingfont\color{sbHead}}{}{0em}{#1}[]
\titleformat{\subsection}{\bfseries\color{sbHead}}{\color{sbHead}\thesubsection.}{0.5em}{#1}[]
\titleformat{\subsubsection}{\bfseries\itshape\color{sbHead}}{\color{sbHead}\thesubsubsection.}{0.5em}{#1}[]
\titleformat{\paragraph}[runin]{\bfseries\color{sbHead}}{}{0em}{#1}

\newcommand{\thead}{\rowcolor{sbTint}\color{sbHead}}
\newcommand{\evt}[2]{\textcolor{#1}{$\bullet$}\,#2}

\newcommand{\tablestyle}[2]{\setlength{\tabcolsep}{#1}\renewcommand{\arraystretch}{#2}}
\makeatletter
\def\@BTrule[#1]{%
  \ifx\longtable\undefined
    \let\@BTswitch\@BTnormal
  \else\ifx\hline\LT@hline
    \nobreak
    \let\@BTswitch\@BLTrule
  \else
     \let\@BTswitch\@BTnormal
  \fi\fi
  \global\@thisrulewidth=#1\relax
  \ifnum\@thisruleclass=\tw@\vskip\@aboverulesep\else
  \ifnum\@lastruleclass=\z@\vskip\@aboverulesep\else
  \ifnum\@lastruleclass=\@ne\vskip\doublerulesep\fi\fi\fi
  \@BTswitch}
\makeatother

\addto\extrasenglish{
}

\renewcommand{\titlefont}{\color{black}\normalfont\bfseries\fontsize{14}{18}\selectfont}

\let\sbmaketitle\maketitle
\renewcommand{\maketitle}{\begingroup\hbadness=10000 \sbmaketitle\par\endgroup}

\reportnumber{} % n/a
\renewcommand{\today}{}

\title{\centering SocietyBench: Forecasting Counterfactual Social-World Evolution}

\newsavebox{\sbfront}
\AtBeginDocument{\savebox{\sbfront}{%
  \setlength{\fboxsep}{6pt}%
  \colorbox{sbTint}{%
  \begin{tabular}{@{}r@{\ \ }l@{}}
  \small\textcolor{sbHead}{\textbf{Project page}} &
    \small\href{https://co-minder.github.io/Societybench/}{\texttt{co-minder.github.io/Societybench}}
  \end{tabular}}%
}}

\author[*]{%
  Zhenran Wang\textsuperscript{*} \quad Zhonghan Bian\textsuperscript{*} \quad Jinsong Li \quad Zhangyang Qi
  \par\vspace{11pt}\centerline{\usebox{\sbfront}}%
}

\begin{abstract}

Large language models (LLMs), and the agents built on top of them, are now benchmarked heavily on whether they can \textit{finish a task}---fix a bug, drive a browser, operate a GUI. A complementary \textbf{social} ability, namely how well a model understands and forecasts the way real social events unfold, has barely been measured. We introduce \textbf{SocietyBench}, an end-to-end benchmark that takes a one-line event topic, collects Web news and social-media posts across five platforms, distills them into a date-indexed timeline that keeps \textit{factual} events and a \textit{public-opinion} layer separate, and then turns every cutoff date on that timeline into an audited bank of forecasting questions. Questions are scored on two orthogonal 100-point axes: \textbf{probability calibration} and \textbf{temporal accuracy}. Before any model sees a timeline, a three-phase procedure replaces every named entity and shifts every date by a per-event constant, turning a real arc into a \textit{counterfactual social world}---structurally identical to what happened, but stripped of the surface labels a model could match against pre-training memory. On five heterogeneous events and $125$ prediction points in Chinese and English editions, the strongest of six frontier LLMs reaches only $75.0$ out of $100$, against a trivial anchor of $50$. The two axes come apart: a model can be calibration-strong but time-weak, or the reverse. Three agent frameworks built on a shared base model fail to improve on that base, and two model-free heuristics trail every LLM. Per-event gaps reach $21.4$ points on a single axis, which is our main argument for evaluating on several events rather than one. All anonymized timelines, question banks, ground truth, and scoring code are released.

\end{abstract}

\begin{document}

\maketitle
\reportnumber{}
\correspondingauthor={Equal contribution.\\
Correspondence: \href{mailto:qizhangyang2000@gmail.com}{qizhangyang2000@gmail.com}, \href{mailto:zhenran.w.1103@gmail.com}{zhenran.w.1103@gmail.com}}

\section{Introduction}
\label{sec:intro}

The past two years produced a wave of LLM and agent benchmarks---SWE-bench, WebArena, OSWorld, $\tau$-bench and their successors---that all ask a version of the same question: \textit{can the model complete a task?} That axis has matured. It also covers only one side of what these systems are used for. The complementary \textit{\textbf{social}} side---how well a model understands and forecasts the dynamics of a real social event---has hardly been quantified at all, and whether current models can do it remains mostly a matter of anecdote.

\begin{figure}[!t]
  \centering
  \includegraphics[width=\textwidth]{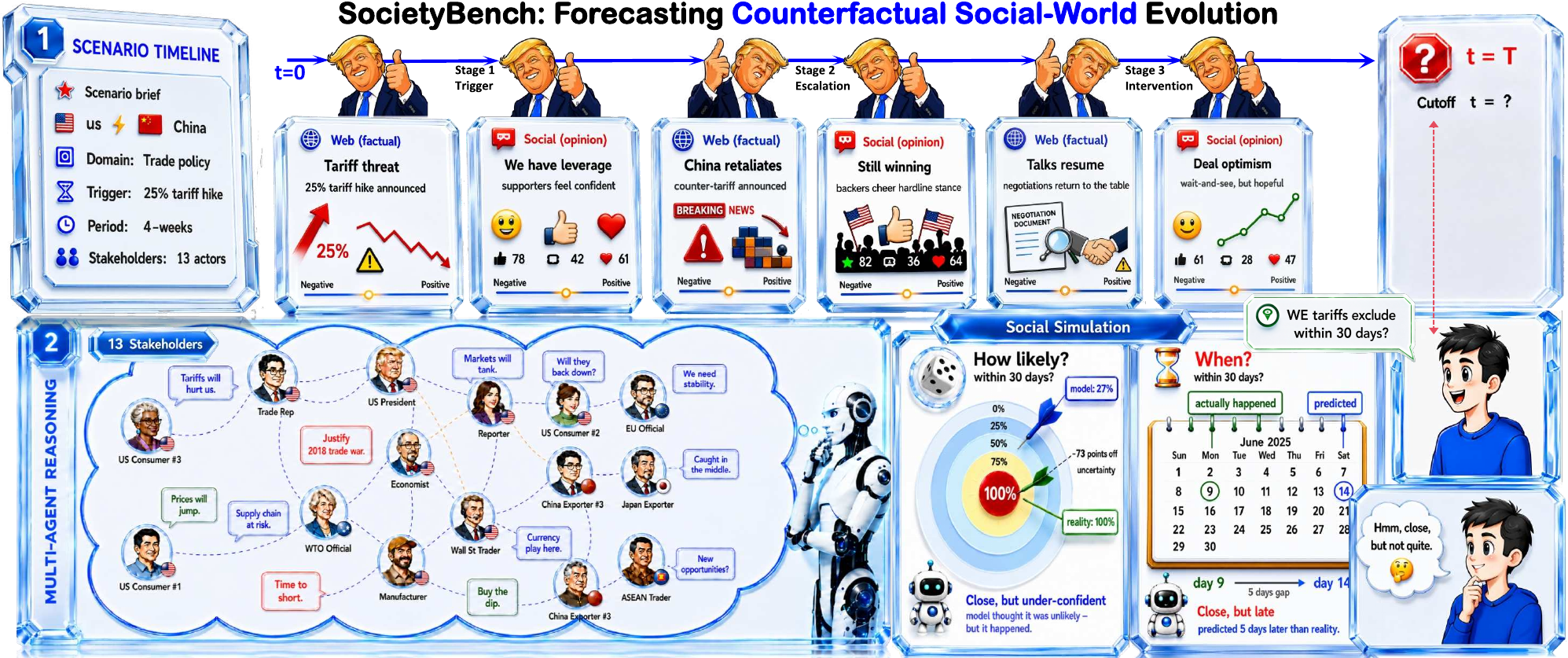}
  \caption{\textbf{Why SocietyBench evaluates counterfactual forecasting rather than event recall.} Evaluated on a real, widely reported event in its original form, a model can score well simply by recognizing the event from pre-training data. SocietyBench replaces the entities and shifts the dates while preserving the causal and temporal structure, so the model has to forecast the future of an arc it cannot recognize.}
  \label{fig:teaser}
\end{figure}

The benchmarks we have are not built to answer it. When a real, widely reported event is evaluated in its original form, a high score can reflect recognition of a memorized arc rather than genuine forecasting (\Cref{fig:teaser} contrasts the two regimes). Autocast~\citep{Autocast} and ForecastBench~\citep{ForecastBench} probe forecasting in a market-style, politico-economic setting, using a curated pool of expert-written ``will $X$ happen by $Y$?'' questions. They are rigorous probability benchmarks, but three gaps keep them out of the regime we care about. First, \textit{the question types are narrow}---nearly every item is a binary yes/no call. Second, \textit{contamination is avoided only by going live}: live questions have to wait on an unresolved future, which forfeits completed arcs, while historical questions already sit in pre-training corpora. Third, \textit{public opinion is ignored}---yet in a social event, narratives flip and sentiment swings faster than the underlying facts change.

A second line of work is LLM-agent social simulation~\citep{GenerativeAgents,AgentSociety,OpinionDynamics,ContextSentiment,FDELLM}: populate a virtual town or forum with LLM agents playing personas, then watch the group behavior that emerges. These works ask whether an LLM can \textit{perform} something that looks like a human society, not whether, given the opening of a real event, it can \textit{predict} what happens next. The first question is about fidelity, the second about forecasting accuracy, and they are different tasks. Because a simulated environment has no real-world ground truth to score against, this line also cannot produce a quantitative leaderboard.

We argue that a benchmark for this regime has to satisfy four demands at once. It must (a) work on \textit{real} events and real multimedia data, so the distribution matches how these systems are actually used; (b) integrate \textit{both} factual reporting and public sentiment, since either alone misses half the dynamics; (c) build its ground truth without expert hand-annotation, so new events can be added as they unfold; and (d) score models with \textit{calibrated} metrics that reward uncertainty-aware predictions rather than accuracy-style hits. No existing benchmark meets all four. In particular (b) and (c) are in tension, because social-media opinion is notoriously hard to align with objective ground truth. Our main design observation is that the conjunction becomes feasible if the benchmark is framed as a \textit{retrospective} forecasting exam on an already-finished event, with the second half of the timeline playing the role of ground truth.

Concretely, we introduce \textbf{SocietyBench}, built on five heterogeneous social events spanning public controversy, geopolitics, technology policy, financial markets, and trade policy. For every event, an automated pipeline distills raw Web news and social-media posts into an anonymized, date-indexed timeline that separates \textit{factual} events from a \textit{public-opinion} layer, and then converts every cutoff into a battery of audited forecasting questions (\Cref{sec:method}). The joint anonymization is more than a leakage filter. By replacing every named entity and shifting every date by a per-event offset, it turns each real arc into a \textit{counterfactual social world}: structurally identical to what really happened, but without the surface labels a model could match against memory. A score therefore reflects \textit{forward reasoning over social-world evolution} rather than recall of the specific real arc the timeline came from.

Our contributions:

\begin{itemize}
  \item We release \textbf{SocietyBench}, a reproducible end-to-end benchmark for LLM and agent forecasting of real social events: five heterogeneous domains, dual-source timelines (Web news plus five social-media platforms), and a fully automated construction pipeline.
  \item We design a benchmarking protocol that pairs input-side \textbf{three-phase entity-and-date anonymization} with output-side \textbf{dual-axis scoring}---probability calibration and temporal accuracy---on calibrated $100$-point metrics.
  \item We evaluate $6$ frontier LLMs, $3$ agent frameworks, and $2$ non-LLM baselines end-to-end on $125$ prediction points in two language editions. The strongest LLM reaches only $75.0$ out of $100$; the two axes dissociate; and per-event gaps reach $21.4$ points on a single axis, which makes single-event evaluation unreliable.
\end{itemize}

\section{Related Work}
\label{sec:related}

\paragraph{Task-completion agent benchmarks.}
Most existing LLM and agent benchmarks target the \textit{task-completion} axis: SWE-bench~\citep{SWEbench} resolves real GitHub issues, WebArena~\citep{WebArena} runs long-horizon web tasks across four self-hosted apps, OSWorld~\citep{OSWorld} covers multi-application desktop workflows, $\tau$-bench~\citep{TauBench} tests policy-bound dialogue agents, and GAIA~\citep{GAIA} poses general-assistant questions requiring tool use. All ask whether an agent can execute a prescribed sequence of operations to reach a \textit{known} end state. SocietyBench is the social-dimension counterpart: instead of asking whether an agent can complete a task whose solution is fixed, we ask whether it can predict how a real-world social event will unfold.

\paragraph{Event and sentiment forecasting benchmarks.}
A separate line poses questions that can only be verified once events play out. Autocast~\citep{Autocast}, retrieval-augmented forecasting~\citep{Halawi2024}, and ForecastBench~\citep{ForecastBench} attach probability questions to dated news corpora. Context-Aware Sentiment Forecasting~\citep{ContextSentiment} narrows to user-sentiment evolution during social-media events. MIRAI~\citep{MIRAI} forecasts country-level geopolitical events from a news database, and FutureX~\citep{FutureX} maintains a daily-refreshed live-web leaderboard. PolyBench~\citep{PolyBench} is the only prior work that scores \textit{time-of-event error in days}, but it does so on Polymarket short-horizon contracts. SocietyBench differs by targeting \textit{medium-to-long-horizon social arcs} that carry both a \textit{factual} and a \textit{public-opinion} layer, rather than short-horizon binary contracts.

\paragraph{LLM-driven social simulation.}
A different line asks the inverse question: rather than \textit{predicting} an event, can we \textit{generate} a plausible society? Generative Agents~\citep{GenerativeAgents} first showed believable sandbox dynamics; AgentSociety~\citep{AgentSociety} and OASIS~\citep{OASIS} scaled the idea to tens of thousands and then a million agents; Opinion Dynamics~\citep{OpinionDynamics} and FDE-LLM~\citep{FDELLM} study how opinions propagate through such networks. A more recent cluster---SocioVerse~\citep{SocioVerse}, MF-LLM~\citep{MFLLM}, DualMind~\citep{DualMind}---moves toward predictive validation, checking whether a simulated population's aggregate trajectory matches a real one. These target aggregate simulator fidelity rather than per-event forecast accuracy. The latter question---given a real unfolding social arc, can an LLM or agent system forecast it?---is what SocietyBench measures.

\section{Methodology}
\label{sec:method}

\subsection{Task Formulation}
\label{sec:method_task}

\begin{figure}[!t]
  \centering
  \includegraphics[width=\textwidth]{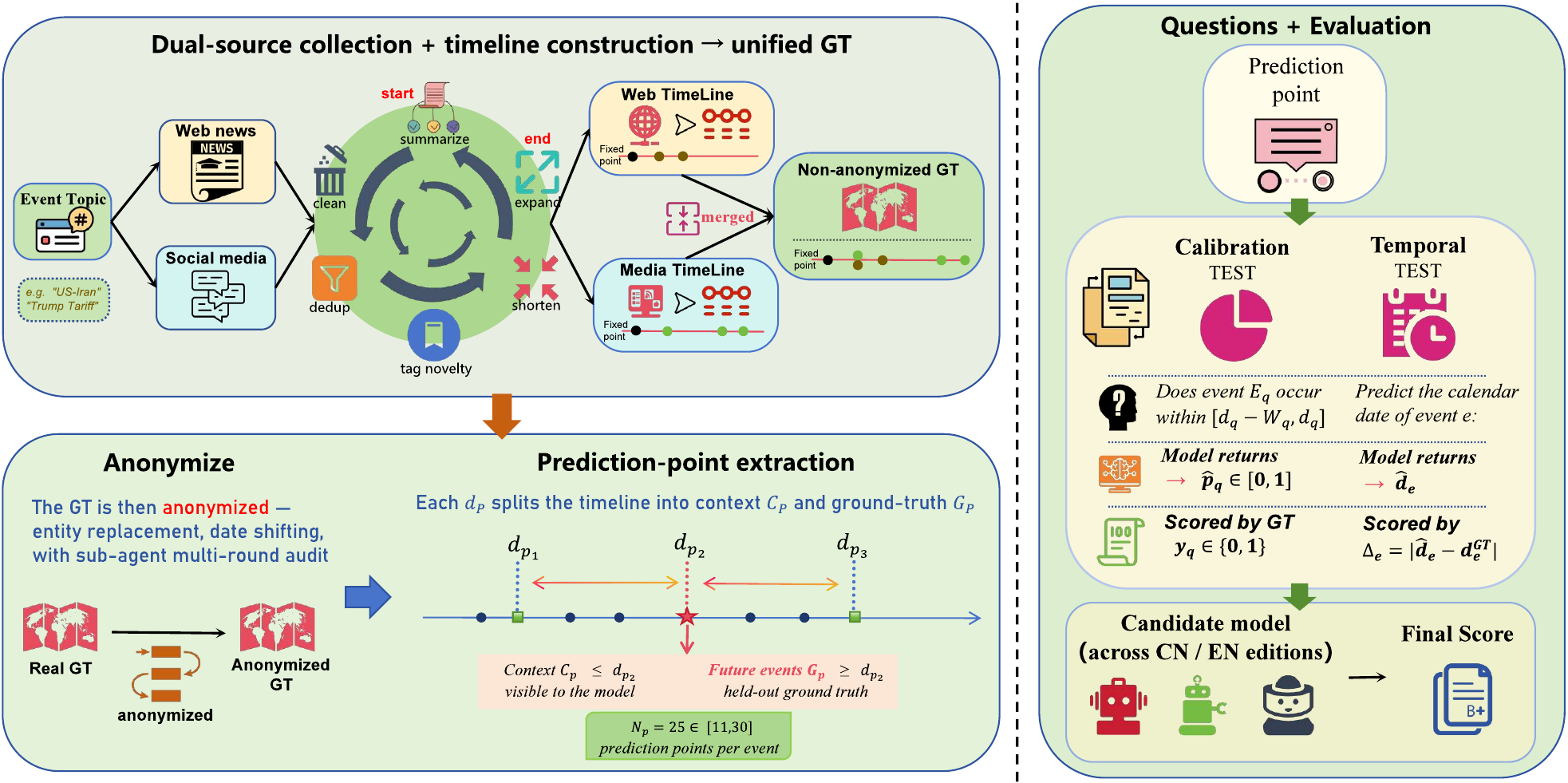}
  \caption{\textbf{SocietyBench methodology.} Each retrospective cutoff $d_P$ splits the anonymized timeline into a visible context $C_P$ and a held-out set $G_P$. An audited bank emits a calibration set $Q^{\text{cal}}_P$ and a temporal set $Q^{\text{time}}_P$, scored on two independent 100-point axes.}
  \label{fig:methodology}
\end{figure}

The model plays a forecaster sitting at a cutoff date $d_P$. It sees everything that happened before $d_P$---the \textit{context} $C_P$---and nothing after. It also cannot tell which real event it is looking at, because every named entity has been replaced and every date shifted (\Cref{sec:method_anon}). It is then asked two kinds of questions about \textit{what happens after} $d_P$ (\Cref{fig:methodology}). Together these form a prediction point
\[
P = \bigl(d_P,\; C_P,\; G_P,\; Q^{\text{cal}}_P,\; Q^{\text{time}}_P\bigr),
\]
where the ground-truth set $G_P$ is simply the timeline nodes from $d_P$ onward.

\begin{itemize}
  \item A \textbf{calibration question} $q = (E_q, d_q, W_q)$ asks whether event $E_q$ occurs within the window $[d_q - W_q,\, d_q]$. The model returns a probability $\hat p_q \in [0,1]$ and is scored against the binary ground truth $y_q \in \{0,1\}$.
  \item A \textbf{temporal question} asks for the calendar date $\hat d_e$ of a future event $e$ in $G_P$. Error is measured in days.
\end{itemize}

Performance is summarized on two orthogonal axes---\textit{probability calibration} and \textit{temporal accuracy}---each converted to a 100-point score per event; full definitions are in \Cref{sec:method_eval}. Every event is evaluated independently, and a bare LLM and a full agent system are run under exactly the same protocol.

\subsection{Dataset and Anonymization}
\label{sec:method_dataset}

The benchmark covers five real events drawn from five domains---\textit{public controversy, geopolitical conflict, technology policy, financial markets, and trade policy}---each spanning at least four weeks, so that the post-cutoff window admits a non-trivial horizon: a \textit{university library harassment dispute} (``Wuhan Library''), a \textit{U.S.--Iran conflict} (``US-Iran''), a \textit{U.S.\ TikTok divestiture/ban ruling} (``TikTok''), a \textit{Super Micro NASDAQ delisting crisis} (``SMCI''), and a \textit{U.S.\ reciprocal-tariff escalation} (``Trump Tariff'').

Each event is processed independently by the same pipeline, which produces a \textit{fact + opinion} merged timeline, a per-event entity-and-date replacement table, an accepted set of prediction points, and a question bank for each prediction point. The pipeline runs automatically end to end; released artifacts additionally pass human acceptance audits, which gate quality but do not hand-author labels. After processing, timeline depth ranges from $39$ to $168$ nodes, with $N_P = 25$ prediction points per event. Per language edition, calibration questions range from $2{,}862$ on \textit{Wuhan Library} to $8{,}035$ on \textit{Trump Tariff}, and declared temporal events from $503$ on \textit{US-Iran} to $976$ on \textit{Trump Tariff}.
\Cref{tab:bank} gives the per-event breakdown.

%% Per-event bank statistics (released banks, per language edition)
\begin{table}[!t]
    \centering
    \tablestyle{6pt}{1.15}
    \resizebox{0.86\textwidth}{!}{%
    \begin{tabular}{lccccc}
        \toprule
        \thead
        \textbf{Event} & \textbf{Nodes} & \textbf{Cal.\ questions} & \textbf{per point} & \textbf{Temporal events} & \textbf{in $0$--$90$\,d window} \\
        \midrule
        Wuhan Library & $39$  & $2{,}862$ & $114.5$ & $534$ & $479$ \\
        US-Iran       & $41$  & $4{,}574$ & $183.0$ & $503$ & $503$ \\
        TikTok        & $127$ & $6{,}697$ & $267.9$ & $571$ & $485$ \\
        SMCI          & $40$  & $3{,}196$ & $127.8$ & $528$ & $381$ \\
        Trump Tariff  & $168$ & $8{,}035$ & $321.4$ & $976$ & $928$ \\
        \midrule
        \textbf{Total} & $\mathbf{415}$ & $\mathbf{25{,}364}$ & $\mathbf{202.9}$ & $\mathbf{3{,}112}$ & $\mathbf{2{,}776}$ \\
        \bottomrule
    \end{tabular}}
    \caption{\textbf{Per-event bank statistics}, per language edition; the two editions are one-to-one. \textit{Nodes} is the timeline depth, and \textit{in-window} counts the temporal events falling $0$--$90$\,d after their cutoff, which is the subset the temporal axis consumes. Every event contributes $N_P = 25$ prediction points.}
    \label{tab:bank}
\end{table}

\paragraph{Anonymization.}
\label{sec:method_anon}
Frontier LLMs may have memorized a raw event---from its named entities down to the absolute calendar position of its arc---so each merged timeline is anonymized before any model sees it, obscuring \textit{named entities} and \textit{dates} jointly. Each event carries its own entity replacement table $\mathcal{R}$ and a single date offset $\delta \sim \mathcal{U}[-\delta_{\max},\, +\delta_{\max}]$ applied uniformly to every date, which destroys absolute calendar position while leaving all inter-date gaps intact. We set $\delta_{\max} = 180$\,d: a window that large crosses two seasons and a full quarterly reporting cycle, which defeats exact temporal lookup, while still keeping seasonal references coherent after the shift.

Two checks bound what memorization can still buy. First, recalling the true calendar cannot pay by construction: re-scoring a model's own predictions as if it had answered on the \textit{real} calendar collapses the temporal score to the same floor that an oracle knowing the true calendar attains against the shifted ground truth. Second, restoring real names and real dates outright moves the calibration axis by only $+1.0$ point on average, so binary outcome judgment does not lean on memorized identities. The three anonymization steps, summarized in \Cref{alg:anon}, are:

\begin{itemize}
  \item $\mathcal{A}$ \textbf{Substitution.} Replace entity strings via $\mathcal{R}$ using longest-match lookup, and shift every date by $\delta$; a final scan repairs doubled substitutions.
  \item $\mathcal{B}$ \textbf{Reverse-identification audit.} A panel of three adversarial LLMs independently re-reads the anonymized text and tries to recover the original event and time period, reporting confidence in $\{\textsc{High}, \textsc{Mid}, \textsc{Low}\}$. \textsc{High} means the auditor named the specific event; \textsc{Mid} means it named the event class; \textsc{Low} means it was wrong. The augment-and-rerun loop follows \Cref{alg:anon}: an event is only released once no auditor still returns \textsc{High}.
  \item $\mathcal{C}$ \textbf{Semantic-consistency pass.} Compare against the original at paragraph, event, and narrative level, repairing any damage introduced by substitution.
\end{itemize}

As argued in \Cref{sec:intro}, the result is a counterfactual social world that a model cannot match against pre-training memory.

\begin{algorithm}[!t]
  \caption{Three-phase anonymization.}
  \label{alg:anon}
  \begin{algorithmic}[1]
    \REQUIRE Raw timeline $\mathcal{T}_{\mathrm{raw}}$, replacement table $\mathcal{R}$, date offset $\delta$, maximum audit rounds $K=5$
    \ENSURE Anonymized timeline $\mathcal{T}_{\mathrm{anon}}$, or event rejection
    \STATE $\mathcal{T}^{(0)} \gets$ replace entities in $\mathcal{T}_{\mathrm{raw}}$ by longest-match lookup in $\mathcal{R}$
    \STATE Shift every date in $\mathcal{T}^{(0)}$ by the same offset $\delta$
    \STATE Repair doubled substitutions and malformed date mentions
    \FOR{$k = 1$ to $K$}
      \STATE Ask the adversarial auditor panel to identify the original event and time window
      \STATE Let $g_k \in \{\textsc{High}, \textsc{Mid}, \textsc{Low}\}$ be the strongest auditor confidence
      \IF{$g_k = \textsc{Low}$}
        \STATE \textbf{break}
      \ELSIF{$g_k = \textsc{Mid}$}
        \STATE Augment $\mathcal{R}$ and/or resample $\delta$ to remove residual clues
        \STATE Re-apply substitution and date shifting
      \ELSE
        \STATE \textbf{return} reject event
      \ENDIF
    \ENDFOR
    \STATE Run paragraph-, event-, and narrative-level semantic-consistency checks
    \STATE Repair any substitution damage while preserving anonymized entities and shifted dates
    \STATE \textbf{return} $\mathcal{T}_{\mathrm{anon}}$
  \end{algorithmic}
\end{algorithm}

\paragraph{Bilingual release.}
Each event's question bank is released in Chinese and English in one-to-one correspondence. Both editions share the same entity placeholders and the same shifted dates, so neither edition can be used to de-anonymize the other.

\subsection{Evaluation Protocol}
\label{sec:method_eval}

Every model receives a \textbf{calibration score} $S_{\text{cal}}$ and a \textbf{temporal score} $S_{\text{time}}$, both on a 0--100 higher-is-better scale.

\paragraph{Calibration score.}
Each calibration question $q$ carries a weight $w_q = w^{\text{time}}_q \cdot w^{\text{win}}_q$, the product of two factors:

\begin{itemize}
  \item \textbf{Time factor} $w^{\text{time}}_q = 1 / (1 + \alpha \cdot \Delta t_q)$ with $\alpha = 0.04$. The further a question resolves from the cutoff, in days $\Delta t_q$, the lower its weight, because far-future questions are noisier.
  \item \textbf{Window factor} $w^{\text{win}}_q = W_q / (W_q + 4)$. The shorter the question window $W_q$, the lower the weight, because very short windows admit only trivial predictions.
\end{itemize}

The weighted mean absolute error is then converted to a 100-point score:
\begin{equation}
  \mathrm{wMAE}_{\text{cal}} \;=\; \frac{\sum_{q} w_q \, \lvert \hat p_q - y_q \rvert}{\sum_{q} w_q},
  \qquad
  S_{\text{cal}} \;=\; 100 \cdot \max\!\bigl(0,\; 1 - \mathrm{wMAE}_{\text{cal}}\bigr).
  \label{eq:wmae}
\end{equation}

A predictor that answers $50\%$ to everything scores exactly $50$, which anchors the scale. The choice of rule is not load-bearing: the scoring ablation of \Cref{sec:ablation} re-scores every answer under proper scoring rules (Brier, log-loss) and the model ordering survives both.

\paragraph{Temporal score.}
For every ground-truth event $e$ the model predicts a date $\hat d_e$, with error $\Delta_e = |\hat d_e - d^{\text{GT}}_e|$ in days. Each event falls in one of three 30-day buckets $[\ell_e, u_e]$---days 0--30, 31--60, or 61--90 after the cutoff---according to its offset $t_e$; day 0 is the cutoff date, which the context strictly precedes. A band of radius $\Delta_e$ centered on the true date, clipped to the bucket, gives the normalized miss:
\begin{equation}
  B_e \;=\; \min(u_e,\, t_e+\Delta_e) - \max(\ell_e,\, t_e-\Delta_e),
  \qquad
  m_e \;=\; \min\!\bigl(1,\; B_e/30\bigr).
  \label{eq:band}
\end{equation}
Reusing the same time factor:
\begin{equation}
  S_{\text{time}} \;=\; 100 \cdot \max\!\Bigl(0,\; 1 - \frac{\sum_e w^{\text{time}}_e\, m_e}{\sum_e w^{\text{time}}_e}\Bigr).
  \label{eq:time}
\end{equation}
A guesser that always answers the midpoint of its bucket scores roughly $50$, so the two axes share a common trivial anchor.

\paragraph{Aggregation.}
\Cref{eq:wmae} and \Cref{eq:time} are computed independently on each event, giving a per-event pair $(S_{\text{cal}}, S_{\text{time}})$. A model's headline score is the \textit{cross-event mean} of the five per-event scores, and per-event scorecards are reported alongside it.

\section{Experiments}
\label{sec:exp}

\subsection{Implementation Details}
\label{sec:exp_setup}

We evaluate six frontier, API-accessible LLMs---\textit{GPT-5.5}, \textit{Gemini-3.5-Flash}, \textit{Claude-Opus-4.8}, \textit{DeepSeek-V4-Pro}, \textit{Kimi-K2.5}, and \textit{Doubao-Seed-2.0-Pro}---together with three agent baselines built on the \textit{Doubao} base model: LangGraph~\citep{LangGraph} (plan-and-solve), AutoGen~\citep{AutoGen} (multi-agent debate), and MiroFish~\citep{MiroFish}, a third-party open-source social-simulation forecaster built on OASIS~\citep{OASIS}. All systems are evaluated on the five events of \Cref{sec:method_dataset}, in both the Chinese and English editions.

All systems are queried deterministically (temperature $0$) under a fixed reasoning budget, with the reasoning trace captured. No system is granted external web search beyond what its framework requires internally. Each system answers both question banks once at every prediction point. A question that still lacks a parseable answer after three retries is excluded from the primary score rather than silently defaulted, so no filled value reaches \Cref{tab:main}. The released materials cover the anonymized timelines, the question banks, and the ground truth; per-question model responses are not part of the release.

\subsection{Main Results}
\label{sec:exp_main}

%% Main leaderboard (Table 1)
\begin{table}[!t]
    \centering
    \tablestyle{2.4pt}{1.2}
    \newcolumntype{N}{>{\centering\arraybackslash}p{8.4mm}}
    \resizebox{\textwidth}{!}{%
    \begin{tabular}{l *{10}{N} !{\hspace{0.5em}} *{2}{N} c}
        \toprule
        \thead
         & \multicolumn{2}{c}{\evt{sbEvA}{\textbf{Wuhan Lib.}}} & \multicolumn{2}{c}{\evt{sbEvB}{\textbf{US-Iran}}} & \multicolumn{2}{c}{\evt{sbEvC}{\textbf{TikTok}}} & \multicolumn{2}{c}{\evt{sbEvD}{\textbf{SMCI}}} & \multicolumn{2}{c}{\evt{sbEvE}{\textbf{Trump Tariff}}} & \multicolumn{3}{c}{\textbf{Avg.}} \\
        \cmidrule(lr){2-3} \cmidrule(lr){4-5} \cmidrule(lr){6-7} \cmidrule(lr){8-9} \cmidrule(lr){10-11} \cmidrule(lr){12-14}
        \thead
        \textbf{System} & Cal. & Time & Cal. & Time & Cal. & Time & Cal. & Time & Cal. & Time & Cal. & Time & Overall \\
        \midrule
        GPT-5.5               & $74.7$ & $55.6$ & $66.3$ & $63.2$ & $78.7$ & $84.4$ & $74.9$ & $84.2$ & $81.7$ & $85.4$ & $\mathbf{75.3}$ & $\mathbf{74.6}$ & $\mathbf{75.0}$ \\
        Gemini-3.5-Flash      & $71.0$ & $59.6$ & $68.9$ & $58.5$ & $79.8$ & $79.4$ & $72.3$ & $73.3$ & $77.8$ & $66.1$ & $\mathbf{73.9}$ & $\mathbf{67.3}$ & $\mathbf{70.6}$ \\
        Claude-Opus-4.8       & $59.0$ & $63.2$ & $55.0$ & $62.3$ & $64.2$ & $72.6$ & $60.8$ & $70.4$ & $73.6$ & $81.7$ & $\mathbf{62.5}$ & $\mathbf{70.0}$ & $\mathbf{66.3}$ \\
        DeepSeek-V4-Pro       & $63.4$ & $58.6$ & $55.9$ & $64.9$ & $65.1$ & $66.7$ & $60.8$ & $65.2$ & $67.5$ & $69.0$ & $\mathbf{62.5}$ & $\mathbf{64.9}$ & $\mathbf{63.7}$ \\
        Kimi-K2.5             & $64.4$ & $59.4$ & $58.4$ & $65.6$ & $62.5$ & $63.9$ & $64.8$ & $64.9$ & $66.4$ & $64.0$ & $\mathbf{63.3}$ & $\mathbf{63.6}$ & $\mathbf{63.4}$ \\
        Doubao-Seed-2.0-Pro   & $65.5$ & $58.3$ & $57.9$ & $60.3$ & $67.0$ & $64.2$ & $60.5$ & $65.0$ & $70.3$ & $65.0$ & $\mathbf{64.3}$ & $\mathbf{62.6}$ & $\mathbf{63.4}$ \\
        \bottomrule
    \end{tabular}}
    \caption{\textbf{Main leaderboard.} Per-event and cross-event calibration / temporal means for six frontier LLMs; each cell averages the Chinese and English editions. A uniform $50\%$ predictor scores exactly $50$ on the calibration axis, and a bucket-midpoint guesser scores about $50$ on the temporal axis.}
    \label{tab:main}
\end{table}

\begin{figure}[!t]
  \centering
  \includegraphics[width=\textwidth]{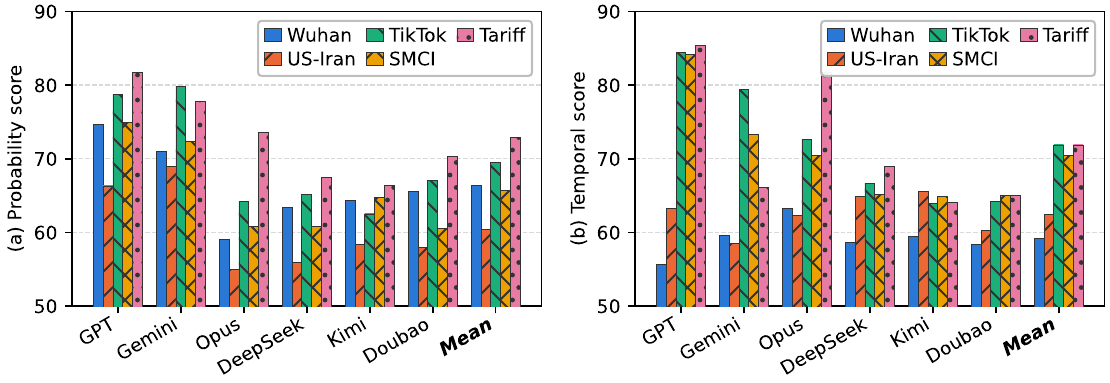}
  \caption{\textbf{Per-event scores.} \textit{Mean} is the six-LLM average on each axis.}
  \label{fig:bars}
\end{figure}

\Cref{tab:main} reports the main leaderboard, with per-event scorecards on the left (visualized in \Cref{fig:bars}) and the cross-event mean on the right; every cell averages the two language editions.

Two findings stand out.

\textbf{The spread is wide at the top and tight at the bottom.} \textit{GPT-5.5} leads at $75.0$, followed by \textit{Gemini-3.5-Flash} at $70.6$ and \textit{Claude-Opus-4.8} at $66.3$, while the bottom three sit within $63.4$--$63.7$ of one another---an $11.3$-point lead ($+17.7\%$) over that cluster. The lead is not an artifact of one lucky event: leaving any single event out and re-averaging still leaves GPT-5.5 ahead by at least $2.4$ points, and it wins outright on four of the five events.

\textbf{The two axes come apart.} \textit{GPT-5.5} leads both ($75.3$ calibration / $74.6$ temporal), but \textit{Gemini-3.5-Flash} is calibration-strong and time-weak ($73.9$ vs.\ $67.3$) while \textit{Claude-Opus-4.8} is the mirror image ($62.5$ vs.\ $70.0$). Neither axis is a proxy for the other, which is the practical reason we report both rather than a single blended number. Even the leader recovers only about half the headroom above the trivial anchor of $50$, so the benchmark is far from saturated.

%% Stress-case comparison (Table 2)
\begin{table}[!t]
    \centering
    \tablestyle{4pt}{1.2}
    \newcolumntype{C}{>{\centering\arraybackslash}p{11mm}}
    \newcolumntype{G}{>{\centering\arraybackslash}p{13mm}}
    \resizebox{\textwidth}{!}{%
    \begin{tabular}{l|CCCC|CC|CC|CC|G@{\hspace{5pt}}G}
        \toprule
        \thead
         & \multicolumn{4}{c|}{\textbf{Event statistics}} & \multicolumn{2}{c|}{\textbf{Mean}} & \multicolumn{2}{c|}{\textbf{GPT-5.5}} & \multicolumn{2}{c|}{\textbf{Kimi-K2.5}} & \multicolumn{2}{c}{\textbf{Gap $\Delta$}} \\
         \cmidrule(lr){2-5} \cmidrule(lr){6-7} \cmidrule(lr){8-9} \cmidrule(lr){10-11} \cmidrule(lr){12-13}
         \thead
         \textbf{Event} & $N_P$ & Arc & \#Cal. & \#Time & Cal. & Time & Cal. & Time & Cal. & Time & Cal. & Time \\
         \midrule
         \textsc{Smci}         & $25$ & $468$d     & $3{,}196$ & $528$ & $65.7$ & $70.5$ & $74.9$ & $84.2$ & $64.8$ & $64.9$ & $\boldsymbol{10.1}_{\text{16\%}}$ & $19.3_{\text{30\%}}$ \\
         \textsc{Trump-Tariff} & $25$ & $2{,}930$d & $8{,}035$ & $976$ & $72.9$ & $71.9$ & $81.7$ & $85.4$ & $66.4$ & $64.0$ & $15.3_{\text{23\%}}$ & $\boldsymbol{21.4}_{\text{33\%}}$ \\
         \bottomrule
    \end{tabular}}
    \caption{\textbf{Stress-case comparison.} \textit{GPT-5.5} (strongest overall) versus \textit{Kimi-K2.5} (weakest on the temporal axis); \textit{Mean} is the six-LLM mean; \textit{Gap $\Delta$} is GPT minus Kimi, with subscripts giving the gap as a percentage of Kimi's score. \textit{Arc} is the span covered by the event's prediction points, and \#Time counts declared events (only those $0$--$90$\,d after a cutoff are scored).}
    \label{tab:stress}
\end{table}

\begin{figure}[!t]
  \centering
  \includegraphics[width=0.94\textwidth]{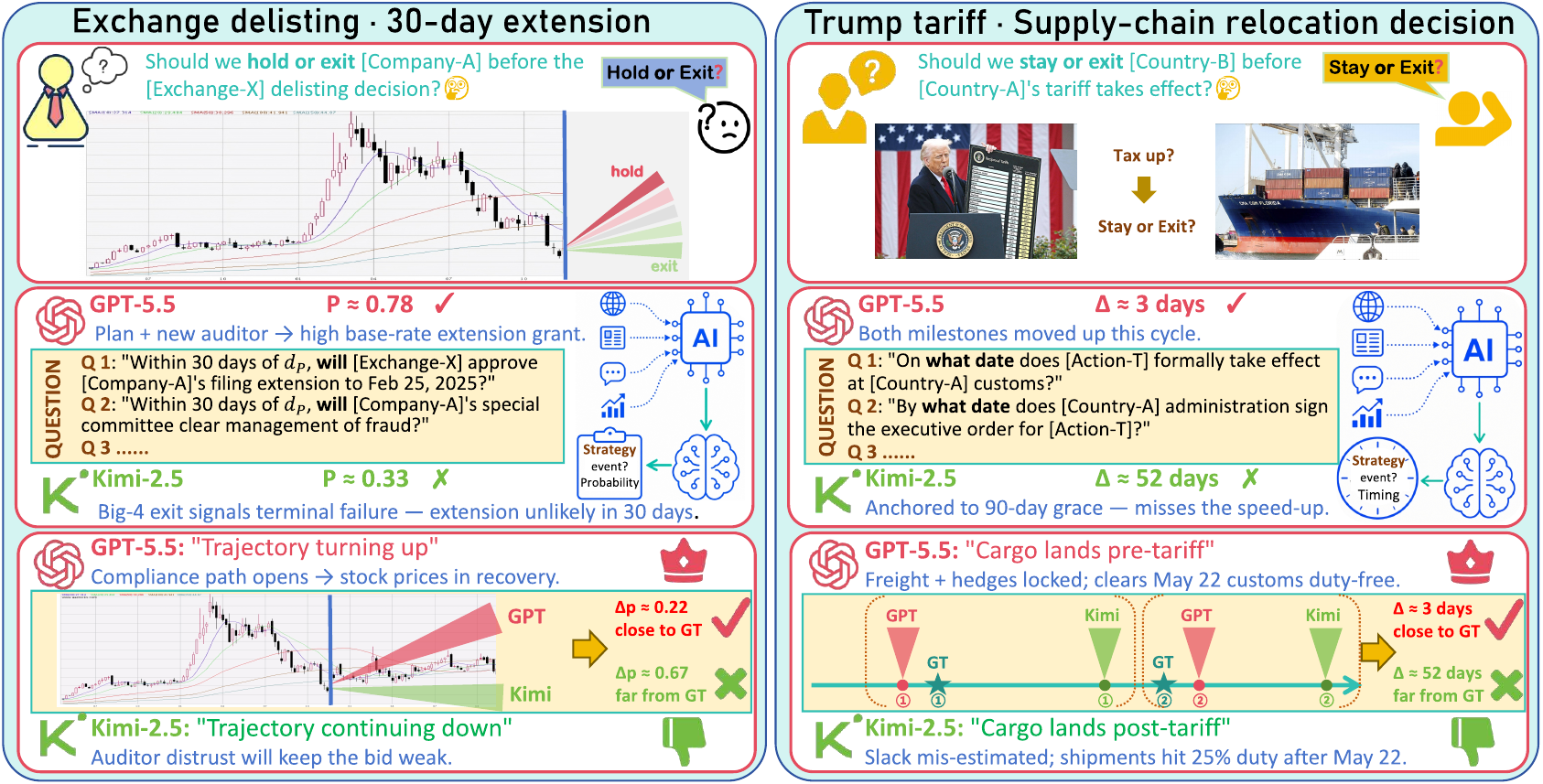}
  \caption{\textbf{Qualitative model comparison.} The \textsc{Smci} extension decision (left) and the tariff decision (right), \textit{GPT-5.5} (\textcolor{sbBoth!70!black}{$\checkmark$}) versus \textit{Kimi-K2.5} (\textcolor{sbTruth}{$\times$}). GPT commits to a calibrated probability and lands close to the true date; Kimi under-commits and anchors far away.}
  \label{fig:qualitative}
\end{figure}

\paragraph{Stress-case events.}
Two events stand out as stress cases (\Cref{tab:stress}; \Cref{fig:qualitative} traces both qualitatively). On \textit{Trump Tariff}, \textit{Kimi-K2.5} scores lowest on both axes ($66.4$ / $64.0$) and the temporal gap to \textit{GPT-5.5} reaches $21.4$ points ($+33\%$)---the largest single-event, single-axis gap in the whole table---on the very event where most models post their best calibration scores. On \textit{SMCI} the same GPT--Kimi gap is double-digit on both axes ($+10.1$ / $+19.3$), which is consistent with an event built around company-specific numerical disclosures that reward precise uncertainty handling. The practical implication is the one we flag in \Cref{sec:intro}: a benchmark of this kind reports a much noisier picture if it is run on a single event---two of our five events would each, on their own, tell a materially different story about how far apart the models are.

\subsection{Ablation Study}
\label{sec:ablation}

We ablate four design choices: question-bank composition, the calibration scoring formula, true/false composition, and cutoff position.

%% Ablation: question-bank composition + calibration scoring formula (Tables 4-5)
\begin{table}[!t]
    \centering
    \begin{minipage}[t]{0.52\linewidth}
        \centering
        {\small
        \tablestyle{3.4pt}{0.98}
        \resizebox{\linewidth}{!}{%
        \begin{tabular}{@{}l|ll|l@{}}
        \toprule[1.5pt]
        \thead
        Question-bank composition & Doubao & GPT & Mean \\ \midrule
        $-$ A-type (time-gradient, 47\%) & $70.7_{+6.4}$ & $83.4_{+8.1}$ & $\mathbf{74.2}_{+7.2}$ \\
        $-$ B-type (specificity, 23\%) & $62.9_{-1.4}$ & $74.1_{-1.2}$ & $\mathbf{66.2}_{-0.8}$ \\
        $-$ C-type (variants, 12\%) & $63.9_{-0.4}$ & $74.8_{-0.5}$ & $\mathbf{66.6}_{-0.4}$ \\
        $-$ D-type (designed-false, 18\%) & $62.1_{-2.2}$ & $72.2_{-3.1}$ & $\mathbf{63.6}_{-3.4}$ \\
        Full bank$_{\text{Cal.}}$ (ours, ${\sim}25$k Q/ed.) & $\mathbf{64.3}$ & $\mathbf{75.3}$ & $\mathbf{67.0}$ \\
        \midrule
        $-$ Near events (0--30\,d, 52\%) & $62.0_{-0.5}$ & $78.0_{+3.4}$ & $\mathbf{67.7}_{+0.5}$ \\
        $-$ Mid events (31--60\,d, 29\%) & $62.8_{+0.3}$ & $74.4_{-0.2}$ & $\mathbf{67.4}_{+0.2}$ \\
        $-$ Far events (61--90\,d, 18\%) & $62.5_{-0.1}$ & $73.5_{-1.1}$ & $\mathbf{66.8}_{-0.4}$ \\
        Full bank$_{\text{Time}}$ (ours, ${\sim}3.1$k ev./ed.) & $\mathbf{62.6}$ & $\mathbf{74.6}$ & $\mathbf{67.2}$ \\
        \bottomrule[1.5pt]
        \end{tabular}}
        }
        \vspace{1mm}
        \caption{\textbf{Question-bank composition.} The upper block drops a question type, the lower block a horizon bucket (shares rounded); subscripts are relative to the full bank.}
        \label{tab:qtype}
    \end{minipage}
    \hfill
    \begin{minipage}[t]{0.44\linewidth}
        \centering
        {\small
        \tablestyle{2.6pt}{0.98}
        \resizebox{\linewidth}{!}{%
        \begin{tabular}{@{}l|ll|l@{}}
        \toprule[1.5pt]
        \thead
        Calibration scoring formula & Doubao & GPT & Gap \\ \midrule
        Equal weights ($w{=}1$) & $64.2_{-0.1}$ & $74.2_{-1.1}$ & $10.0$ \\
        Linear time-decay ($1{-}d/365$) & $65.4_{+1.1}$ & $76.0_{+0.7}$ & $10.6$ \\
        No time-decay ($\alpha{=}0$) & $64.8_{+0.5}$ & $75.0_{-0.3}$ & $10.2$ \\
        Steeper decay ($\alpha{=}0.08$) & $63.8_{-0.5}$ & $75.1_{-0.2}$ & $11.3$ \\
        Steeper decay ($\alpha{=}0.16$) & $63.1_{-1.2}$ & $75.0_{-0.3}$ & $11.9$ \\
        Brier score (rescaled) & $57.4_{-6.9}$ & $70.8_{-4.5}$ & $13.4$ \\
        Log-loss (rescaled) & $45.4_{-18.9}$ & $65.0_{-10.3}$ & $19.6$ \\
        Hyperbolic score mapping & $73.4_{+9.1}$ & $80.1_{+4.8}$ & $6.7$ \\
        Weighted MAE (ours, $\alpha{=}0.04$) & $\mathbf{64.3}$ & $\mathbf{75.3}$ & $\mathbf{11.0}$ \\
        \bottomrule[1.5pt]
        \end{tabular}}
        }
        \vspace{1mm}
        \caption{\textbf{Calibration scoring formula.} Alternative weights, metrics, and mappings; \textit{Gap} is GPT minus Doubao.}
        \label{tab:scoring}
    \end{minipage}
\end{table}

%% Ablation: true/false bias + cutoff gradient (Tables 6-7)
\begin{table}[!t]
    \centering
    \begin{minipage}[t]{0.47\linewidth}
        \centering
        {\small
        \tablestyle{2.2pt}{1.0}
        \resizebox{\linewidth}{!}{%
        \begin{tabular}{@{}l|*{5}{c}|c@{}}
        \toprule[1.5pt]
        \thead
        True/false Q's & Doubao & DSeek & Opus & Gemini & GPT & Mean (6) \\ \midrule
        True  & $59.8$ & $49.3$ & $46.4$ & $72.0$ & $73.5$ & $\mathbf{60.3}$ \\
        False & $71.6$ & $79.9$ & $84.3$ & $76.6$ & $78.3$ & $\mathbf{76.1}$ \\
        Bias  & ${\scriptstyle+}\mathbf{11.8}$ & ${\scriptstyle+}\mathbf{30.6}$ & ${\scriptstyle+}\mathbf{38.0}$ & ${\scriptstyle+}\mathbf{4.7}$ & ${\scriptstyle+}\mathbf{4.8}$ & ${\scriptstyle+}\mathbf{15.9}$ \\
        \bottomrule[1.5pt]
        \end{tabular}}
        }
        \vspace{1mm}
        \caption{\textbf{True/false-question bias (calibration axis).} Models deny false events better than they confirm true ones; \textit{Bias} is false minus true.}
        \label{tab:bias}
    \end{minipage}
    \hfill
    \begin{minipage}[t]{0.49\linewidth}
        \centering
        {\small
        \tablestyle{2.4pt}{0.98}
        \resizebox{\linewidth}{!}{%
        \begin{tabular}{@{}l|*{4}{c}|c@{}}
        \toprule[1.5pt]
        \thead
        Cutoff position & Doubao & Opus & Gemini & GPT & Mean (6) \\ \midrule
        Early  & $63.4$ & $62.7$ & $75.1$ & $75.5$ & $\mathbf{66.7}$ \\
        Middle & $66.1_{+2.7}$ & $63.5_{+0.8}$ & $73.6_{-1.5}$ & $75.4_{-0.1}$ & $\mathbf{67.6}_{+0.9}$ \\
        Late   & $66.6_{+0.5}$ & $68.4_{+4.9}$ & $69.6_{-4.0}$ & $77.4_{+2.0}$ & $\mathbf{68.7}_{+1.1}$ \\
        Gradient & ${\scriptstyle+}\mathbf{3.2}$ & ${\scriptstyle+}\mathbf{5.7}$ & ${\scriptstyle-}\mathbf{5.5}$ & ${\scriptstyle+}\mathbf{1.9}$ & ${\scriptstyle+}\mathbf{2.0}$ \\
        \bottomrule[1.5pt]
        \end{tabular}}
        }
        \vspace{1mm}
        \caption{\textbf{Cutoff gradient (calibration axis).} Later prediction points see more history and score higher; \textit{Gradient} is late minus early.}
        \label{tab:cutoff}
    \end{minipage}
\end{table}

\paragraph{Question-bank composition.}
\Cref{tab:qtype} drops one subset of the bank at a time and re-scores. The four audited types are: \textbf{A}, does the same event occur within windows growing from $7$ to $90$ days; \textbf{B}, one fact re-asked at rising levels of detail; \textbf{C}, true outcomes paired with counterfactual variants; \textbf{D}, events built to be false. Dropping A inflates the six-LLM mean the most ($+7.2$), so A carries most of the benchmark's difficulty. It targets a specific failure: models over-bet on a currently hot narrative continuing in the near term, which is exactly what a growing-window question probes. Dropping D moves the mean the other way ($-3.4$), making it the easiest subset---models are good at rejecting invented events. B and C shift the mean by at most $0.8$, so neither changes the ranking on its own. On the temporal side, dropping any horizon bucket moves the mean by at most $0.5$, so that axis is robust to the horizon mix.

\paragraph{Calibration scoring formula.}
\Cref{tab:scoring} re-scores the same answers under eight alternative calibration rules. Removing the time decay, steepening it, or dropping the question weights altogether moves scores by at most $1.2$ points, so the score is insensitive to the exact weighting. Replacing MAE with Brier or log-loss deflates Doubao by $6.9$ and $18.9$ points respectively; a hyperbolic mapping inflates both models and shrinks their gap to $6.7$. Under every rule the GPT--Doubao gap stays positive, so the ranking does not depend on the choice of rule. And against the suspicion that a designer picks the metric that flatters the result: the proper scoring rules (Brier, log-loss) \textit{widen} the gap rather than narrowing it, because they punish confident errors harder and weaker models make more of them. Our weighted MAE is the gentler choice. We keep it because it sits in the middle of the score range while up-weighting the near-cutoff questions, where forecasting is hardest, and down-weighting trivially short windows.

\paragraph{True/false-question bias.}
\Cref{tab:bias} splits the calibration bank by answer: events that did occur, versus the false variants of types A--C and the designed-false D set. Models are much better at \textit{denying false events} than at \textit{confirming true ones}---six-LLM means of $76.1$ against $60.3$, and a mean per-model bias of $+15.9$. This is a disposition, not raw strength. The two leaders are near balanced (\textit{GPT-5.5} $+4.8$, \textit{Gemini-3.5-Flash} $+4.7$); \textit{Claude-Opus-4.8} is the extreme at $+38.0$, very good at rejecting what did not happen and poor at endorsing what did. A different true/false ratio would reward that disposition, so we keep the bank near $1{:}1$---the axis should measure forecasting, not default skepticism.

\paragraph{Cutoff gradient.}
\Cref{tab:cutoff} splits each event's $25$ prediction points into early, middle, and late terciles. Later cutoffs, which see more of the arc, score higher: $66.7 \to 67.6 \to 68.7$. This is the ``closer is easier'' property a forecasting benchmark should show, and a basic sanity check---flat or inverted scores would mean the questions are not measuring forecasting from context. The trend holds for five of the six LLMs, reaching $+5.7$ on \textit{Claude-Opus-4.8}. \textit{Gemini-3.5-Flash} is the exception at $-5.5$, best at the earliest cutoffs: a quirk of that model, not of the benchmark, and worth knowing before reading any single-cutoff evaluation of it.

\subsection{Agents and Non-LLM Baselines}
\label{sec:exp_agents}

Two questions put the leaderboard in context: does agent orchestration lift a base LLM, and how much of the LLMs' performance can a model-free heuristic recover?

\paragraph{Agent orchestration.}
\Cref{tab:agents} runs three orchestration frameworks on the \textit{Doubao} base. None improves on the base: LangGraph loses $1.3$ points overall, AutoGen $2.9$, and MiroFish $0.3$. The losses concentrate on the calibration axis, reaching $-5.8$ for AutoGen, while temporal scores stay within $0.3$ of the base---orchestration perturbs probabilities far more than it perturbs dates. Read against \Cref{tab:main}, the base model, not the scaffold, determines where a system lands; no agent setup approaches the leading single LLMs. This is a negative result worth stating plainly, because the intuition that a debate or plan-and-solve wrapper should help on a reasoning-heavy forecasting task is a natural one, and it does not hold here.

%% Agent orchestration (Table 3)
\begin{table}[!t]
    \centering
    \tablestyle{2.4pt}{1.2}
    \newcolumntype{N}{>{\centering\arraybackslash}p{8.4mm}}
    \resizebox{\textwidth}{!}{%
    \begin{tabular}{l *{10}{N} !{\hspace{0.5em}} ccc}
        \toprule
        \thead
         & \multicolumn{2}{c}{\evt{sbEvA}{\textbf{Wuhan Lib.}}} & \multicolumn{2}{c}{\evt{sbEvB}{\textbf{US-Iran}}} & \multicolumn{2}{c}{\evt{sbEvC}{\textbf{TikTok}}} & \multicolumn{2}{c}{\evt{sbEvD}{\textbf{SMCI}}} & \multicolumn{2}{c}{\evt{sbEvE}{\textbf{Trump Tariff}}} & \multicolumn{3}{c}{\textbf{Avg.}} \\
        \cmidrule(lr){2-3} \cmidrule(lr){4-5} \cmidrule(lr){6-7} \cmidrule(lr){8-9} \cmidrule(lr){10-11} \cmidrule(lr){12-14}
        \thead
        \textbf{System} & Cal. & Time & Cal. & Time & Cal. & Time & Cal. & Time & Cal. & Time & Cal. & Time & Overall \\
        \midrule
        Doubao (base) & $65.5$ & $58.3$ & $57.9$ & $60.3$ & $67.0$ & $64.2$ & $60.5$ & $65.0$ & $70.3$ & $65.0$ & $\mathbf{64.3}$ & $\mathbf{62.6}$ & $\mathbf{63.4}$ \\
        \quad LangGraph & $62.7$ & $55.8$ & $55.9$ & $60.4$ & $65.8$ & $64.7$ & $56.6$ & $65.5$ & $69.0$ & $64.9$ & $\mathbf{62.0}_{-2.3}$ & $\mathbf{62.3}_{-0.3}$ & $\mathbf{62.1}_{-1.3}$ \\
        \quad AutoGen & $59.2$ & $57.2$ & $52.6$ & $61.4$ & $63.9$ & $64.7$ & $50.3$ & $64.5$ & $66.6$ & $65.0$ & $\mathbf{58.5}_{-5.8}$ & $\mathbf{62.5}_{-0.1}$ & $\mathbf{60.5}_{-2.9}$ \\
        \quad MiroFish & $66.2$ & $57.5$ & $57.1$ & $59.2$ & $65.5$ & $65.5$ & $59.2$ & $65.7$ & $70.3$ & $64.4$ & $\mathbf{63.7}_{-0.6}$ & $\mathbf{62.4}_{-0.2}$ & $\mathbf{63.1}_{-0.3}$ \\
        \bottomrule
    \end{tabular}}
    \caption{\textbf{Agent orchestration.} Three agent frameworks running on the \textit{Doubao} base model (first row); subscripts give the change relative to the base.}
    \label{tab:agents}
\end{table}

\paragraph{Non-LLM baselines.}
Two model-free heuristics anchor the scale. A \textit{frequency baseline} answers each calibration question with the empirical base rate of similar events in the visible context, and a \textit{momentum baseline} does the same over a 7-day rolling window, implementing exactly the near-term-continuation assumption that the ablation above identifies as the dominant LLM failure mode. They score $53.8$ and $54.2$ on the calibration axis---$8$ to $21$ points below every LLM in \Cref{tab:main}, and only about $4$ points above the trivial anchor. Neither emits dates, so their temporal score is recorded at the bucket-midpoint anchor of $50$. Two things follow: the leaderboard reflects predictive signal rather than an artifact of the scoring formula, and the failure mode the LLMs fall into carries almost no predictive value as a deliberate strategy.

\section{Conclusion}
\label{sec:conclusion}

We presented \textbf{SocietyBench}, an end-to-end benchmark that anonymizes real social events into \textit{counterfactual social worlds}---arcs that can be predicted only by genuine social-dynamic reasoning, not by pre-training recall---and scores models on probability calibration and temporal accuracy. Across five heterogeneous events and $125$ prediction points in two language editions, the strongest of six LLMs reaches only $75.0$ against a trivial anchor of $50$; three agent frameworks fail to beat their base LLM; and two non-LLM baselines trail every LLM. Per-event gaps reach $21.4$ points on a single axis, which is our main argument for multi-event over single-event evaluation.

Useful next steps are stronger anonymization to separate forecasting from memory even more cleanly, a broader event pool, finer discrimination among near-peer models, and community-contributed events and systems.

\bibliographystyle{abbrvnat}
\bibliography{main}

\end{document}